\documentclass[10pt]{IEEEtran}
\usepackage{cite}
\usepackage{amsmath,amssymb,amsfonts}
\usepackage{graphicx}
\usepackage{textcomp}
\usepackage{lipsum, color}
\usepackage{graphics}
\newcommand{\subparagraph}{}
\usepackage{titlesec}
\usepackage{subfig}
\usepackage{graphicx}
\usepackage{xcolor}
\usepackage{titlesec}
\usepackage{subfig}
\usepackage{graphicx}
\usepackage{mathtools}
\usepackage{siunitx}
\usepackage{soul}
\usepackage{balance}
\usepackage{algpseudocode}
\usepackage{cite}
\usepackage[font=footnotesize]{caption}
\usepackage{tabularx}
\usepackage{array}
\usepackage{floatrow}
\usepackage[english]{babel}
\usepackage[utf8x]{inputenc}
\usepackage{amsmath}
\usepackage{xcolor}
\usepackage{tikz}
\usepackage{blindtext}
\usepackage{enumitem}
\usepackage[T1]{fontenc}
\usepackage{flushend}
\IEEEoverridecommandlockouts

\def\BibTeX{{\rm B\kern-.05em{\sc i\kern-.025em b}\kern-.08em
    T\kern-.1667em\lower.7ex\hbox{E}\kern-.125emX}}

\begin{document}
\title{MC-CIM: Compute-in-Memory with Monte-Carlo Dropouts for Bayesian Edge Intelligence}

\author{Priyesh Shukla, ~\IEEEmembership{Graduate Student Member,~IEEE,} Shamma Nasrin, ~\IEEEmembership{Student Member,~IEEE,} Nastaran Darabi, ~\IEEEmembership{Student Member,~IEEE,} Wilfred Gomes, and Amit Ranjan Trivedi, ~\IEEEmembership{Senior Member,~IEEE}
\thanks{Priyesh Shukla, Shamma Nasrin, Nastaran Darabi, and Amit Ranjan Trivedi are with the Department of Electrical and Computer Engineering, University of Illinois at Chicago, IL 60607 USA (e-mail: pshukl23@uic.edu; snasri2@uic.edu; ndarab2@uic.edu; amitrt@uic.edu). 

Wilfred Gomes is with Intel, Santa Clara, CA 95054 USA (e-mail:
wilfred.gomes@intel.com).}

} 

\maketitle

\begin{abstract}

We propose MC-CIM, a compute-in-memory (CIM) framework for robust, yet low power, Bayesian edge intelligence. Deep neural networks (DNN) with deterministic weights cannot express their prediction uncertainties, thereby pose critical risks for applications where the consequences of mispredictions are fatal such as surgical robotics. To address this limitation, Bayesian inference of a DNN has gained attention. Using Bayesian inference, not only the prediction itself, but the prediction confidence can also be extracted for planning risk-aware actions. However, Bayesian inference of a DNN is computationally expensive, ill-suited for real-time and/or edge deployment. An approximation to Bayesian DNN using Monte Carlo Dropout (MC-Dropout) has shown high robustness along with low computational complexity. Enhancing the computational efficiency of the method, we discuss a novel CIM module that can perform in-memory probabilistic dropout in addition to in-memory weight-input scalar product to support the method. We also propose a \textit{compute-reuse} reformulation of MC-Dropout where each successive instance can utilize the product-sum computations from the previous iteration. Even more, we discuss how the random instances can be optimally ordered to minimize the overall MC-Dropout workload by exploiting combinatorial optimization methods. Application of the proposed CIM-based MC-Dropout execution is discussed for MNIST character recognition and visual odometry (VO) of autonomous drones. The framework reliably gives prediction confidence amidst non-idealities imposed by MC-CIM to a good extent. Proposed MC-CIM with 16$\times$31 SRAM array, 0.85 V supply, 16nm low-standby power (LSTP) technology consumes 27.8 pJ for 30 MC-Dropout instances of probabilistic inference in its most optimal computing and peripheral configuration, saving $\sim$43\% energy compared to typical execution. 
\end{abstract}

\section{Introduction}
Bayesian inference of deep neural networks (DNNs) can enable higher predictive robustness in decision-making \cite{neal2012bayesian, ghahramani2015probabilistic}. Unlike classical inference where the network parameters such as layer-weights are learned deterministically, Bayesian inference learns them statistically to express model's uncertainty along with the prediction itself. For many edge devices such as insect-scale drones \cite{chukewad2021robofly} and augmented/virtual reality (AR/VR) glasses \cite{rambach2021survey}, such Bayesian inference of DNNs is critical since the application spaces are highly dynamic whereas the consequences of mispredictions can be fateful. Using Bayesian inference, prediction confidence can be systematically accounted in decision making and risk-prone actions can be averted when the prediction confidence is low. 

Nonetheless, Bayesian inference of deep learning models is also considerably more demanding than classical inference. To reduce the computational workload of Bayesian inference, efficient approximations have been explored. For example, Variational inference reduces the learning and inference complexities of fully-fledged Bayesian inference by approximating weight uncertainties using parametric distributions (such as Gaussian Mixture Models). Therefore, with Variational inference, only the parameters of the approximating statistical distributions are learned. Inference workload also simplifies by operating with analytically-defined density models, rather than Markov Chain Monte-Carlo (MCMC)-based procedures as in true Bayesian inference \cite{neal2012bayesian}. 

Even with Variational approximations, the workload of Bayesian inference remains formidable for area/power-constrained devices such as nano-drones and AR/VR glasses. To further simplify the complexities, in \cite{gal2016dropout}, a dropout-based inference procedure, Monte Carlo dropout (MC-Dropout), was developed where dropout used during training is also used for inference. Predictions from many dropout iterations of deep learning model are averaged to determine the net output, whereas the variance estimates the prediction confidence. In \cite{gal2016dropout}, Gal et al. demonstrated that such dropout-based inference is, in fact, an efficient Variational approximation of true Bayesian inference. Robustness of such dropout-based Variational inference has been demonstrated for applications such as character recognition \cite{gal2016dropout}, pose estimation of an autonomous drone \cite{kendall2016modelling}, and RNA sequencing \cite{breda2021bayesian}. 

\begin{figure*}[t]
    \centering
    \includegraphics[width=0.9\linewidth]{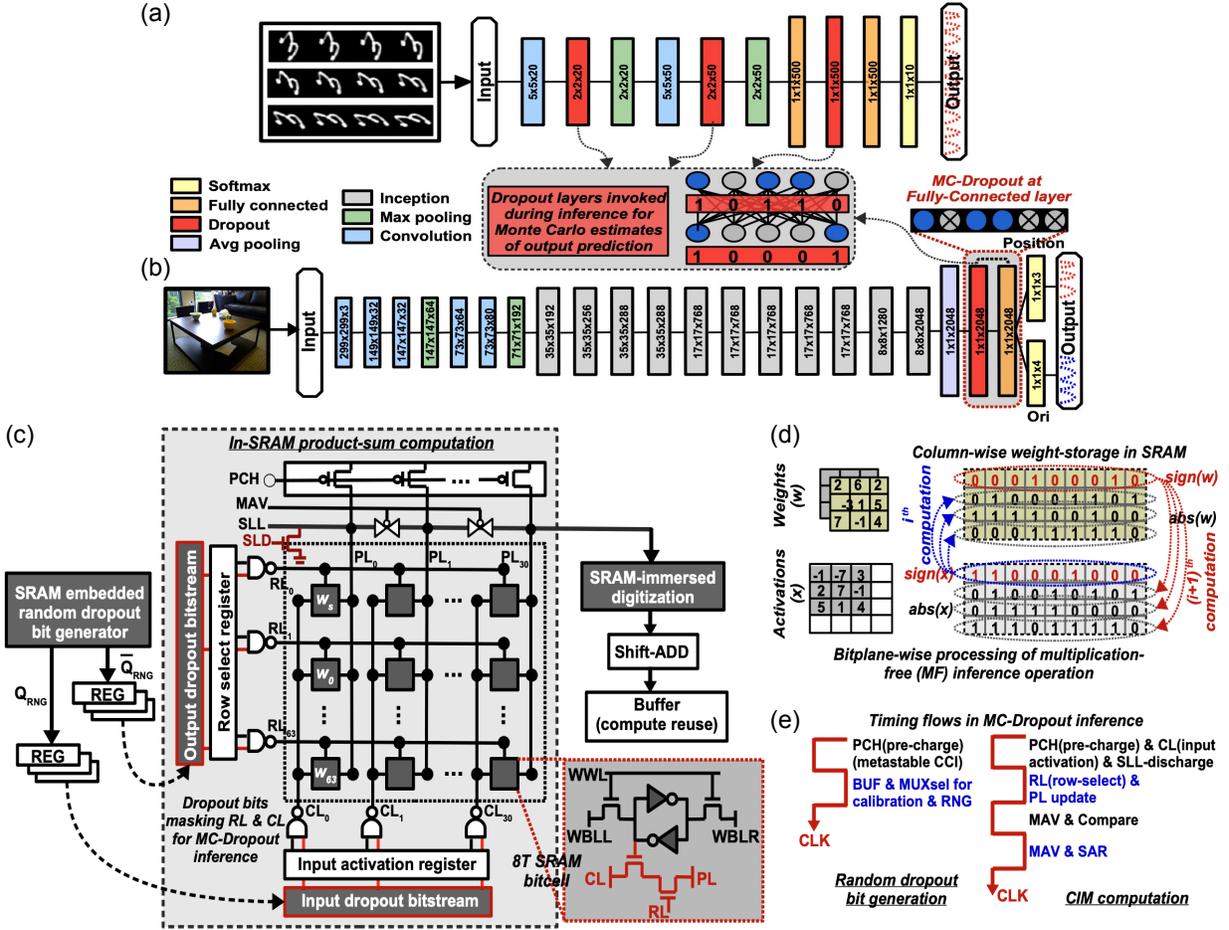}
    \caption{Compute-in-Memory (CIM) framework for efficient acceleration of Monte-Carlo Dropout (MC-Dropout)-based Bayesian inference (BI). (a) LeNet-5 with intermediate dropout layers for uncertainty-aware handwritten digit recognition. (b) Modified Inception v3 for uncertainty-aware visual odometry. (c) Static random access memory (SRAM)-based CIM macro integrating storage and Bayesian inference (BI). Inset figure: 8T SRAM cell with storage and product ports. CIM is embedded with random dropout bit generator for MC-Dropout inference. (d) Bitplane-wise processing of multiplication-free (MF) operator in SRAM-CIM removing digital-to-analog converter (DAC) overhead. (e) Timing flows of CIM computation primitives.}
    \label{fig:syslevel}
\vspace{-1em}
\end{figure*}

In this work, we harness the predictive robustness of MC-Dropout-based Variational inference for \textit{robust edge intelligence} using MC-CIM. Towards this, our key contributions are:

\begin{itemize}[leftmargin=3mm,itemsep=1pt,topsep=0pt,parsep=0pt,partopsep=0pt]
\item We present low-complexity, static random access memory (SRAM)-based compute-in-memory (CIM) macros termed as MC-CIM to accelerate probabilistic iterations of MC-Dropout. Using MC-CIM, the prediction confidence can be extracted along with the prediction itself. MC-CIM also exploits \textit{co-designed} compute-in-memory inference operators; thereby, unlike \cite{zhang2017memory, shukla2020mc, biswas2018conv}, it does not require digital-to-analog converters (DAC) even for multibit precision operations. SRAM-immersed asymmetric search analog-to-digital converter (ADC) is used where product-sum-statistics is exploited for time-efficiency, i.e., to minimize the necessary conversion cycles. For probabilistic input/output activations, we discuss in-memory random number generators that exploit parasitic leakage current and bitline capacitance for low overhead and programmable sampling. 

\item To minimize the computing workload of MC-CIM, we discuss a novel approach of \textit{compute reuse} between consecutive iterations. In this approach, the output of each iteration is expressed as a function of the output from the previous. Thereby, using such nested evaluations, the necessary workload for each iteration minimizes. Even more, we discuss that probabilistic iterations of MC-Dropout can be optimally ordered to maximize compute reuse opportunities. In particular, we discuss a combinatorial optimization framework based on travelling salesman problem (TSP) to demonstrate such optimal sample ordering. Our evaluations show that with compute reuse and optimal sample ordering, the necessary workload for many probabilistic iterations of MC-Dropout only increases marginally.         

\item We discuss the synergy of compute-in-memory and Variational inference in MC-CIM. We find that MC-Dropout-based inference flow can reduce the necessary network size and precision to match the prediction accuracy of a comparable deterministic inference. Therefore, MC-Dropout and CIM benefit each other where MC-Dropout reduces the necessary network size to alleviate the storage bottleneck of CIM under area constraints. Whereas MC-CIM executes each iteration of the method with much higher energy efficiency to alleviate the workload.

\item We discuss the application of MC-CIM for confidence aware predictions in character recognition and visual odometry (VO) in autonomous insect-scale drones. Especially, while there is a growing interest in real-time edge-AI, many applications can be highly vulnerable to misprediction errors. Lightweight Bayesian processing of MC-CIM is highly suited by extracting both the prediction and prediction confidence within limited area/power bounds. Especially, through these applications, we also discuss the interaction of hardware non-idealities and the fidelity of Bayesian DNNs.

\end{itemize}

In section II we introduce CIM macros with co-optimized inference operator. In section III we present MC-CIM for in-memory MC-Dropout (probabilistic) inference. In section IV we discuss our dataflow optimization schemes. Power-performance characterization of MC-CIM is discussed in section V. In section VI we perform confidence-aware inference for two benchmark applications and conclude in section VII.

\section{SRAM-based Compute-in-Memory Macro with Co-designed Learning Operators}
\subsection{Compute-in-Memory Optimized Inference Operator}
In our prior works \cite{nasrin2021mf, nasrin2021compute}, we showed that the complexity of CIM-based deep learning inference can be significantly simplified by \textit{co-designing} the learning operator against CIM's physical and operating constraints. The novel operator from our prior work correlates weight $\mathbf{w}$ and input feature $\mathbf{x}$ as

\begin{equation} \label{eq:1}
    \mathbf{w} \oplus \mathbf{x} = \sum_i \text{sign}(x_i)\cdot \text{abs}(w_i) + \text{sign}(w_i)\cdot \text{abs}(x_i)
\end{equation}
Here, $\sum$ performs vector sum, $\text{sign}()$ is signum function operator extracting the sign $\pm 1$, $\cdot$ performs element-wise multiplication, $\text{abs}()$ produces absolute value of the operand and $+$ is element-wise addition operator. Note that unlike the typical deep learning operator, $\mathbf{w} \cdot \mathbf{x}$, where multibit weight and input vectors are multiplied, the novel operator multiplies one-bit $\text{sign}(\mathbf{x})$ against multibit abs($\mathbf{w}$), and one-bit $\text{sign}(\mathbf{w})$ against multibit abs($\mathbf{x}$). Such decoupling of multibit operands benefits CIM since using a digital bit-plane-wise operation, i.e., operating on a digital plane of same significance bits in $\mathbf{w}$ and $\mathbf{x}$ in one cycle, can be pursued and digital-to-analog converters can be eliminated. Figure \ref{fig:syslevel}(d) shows such bitplane-wise digital processing of the operator. Importantly, if a similar bitplane-wise processing is followed for the conventional operator, the total number of cycles grow as $n^2$ for $n$-bit precision whereas for our operator, they grow as $2(n-1)$.

\subsection{Compute-in-Memory (CIM) Macro Architecture}
Figure \ref{fig:syslevel}(c) shows the baseline CIM macro architecture using eight transistor static random access memory (8T-SRAM). The inset in Figure \ref{fig:syslevel}(c) shows an 8T-SRAM cell with various access ports for write and CIM operations. The write word line (WWL) selects a cell for write operation and the data bit is written through the left and right write bit lines (WBLL and WBLR). During inference, input bit is applied to cell using the column-line (CL) port and output is evaluated on the product-line (PL). The row line (RL) connects the bitcells horizontally to select weight bits in the respective row for within-memory inference. The CIM array operates in a bitplane-wise manner directly on the digital inputs to avoid digital-to-analog converters (DACs). Bitplane of like-significance input and weight vectors are processed in one cycle as shown in Figure \ref{fig:syslevel}(d).

\begin{figure}[t]
    \centering
    \includegraphics[width=0.9\columnwidth]{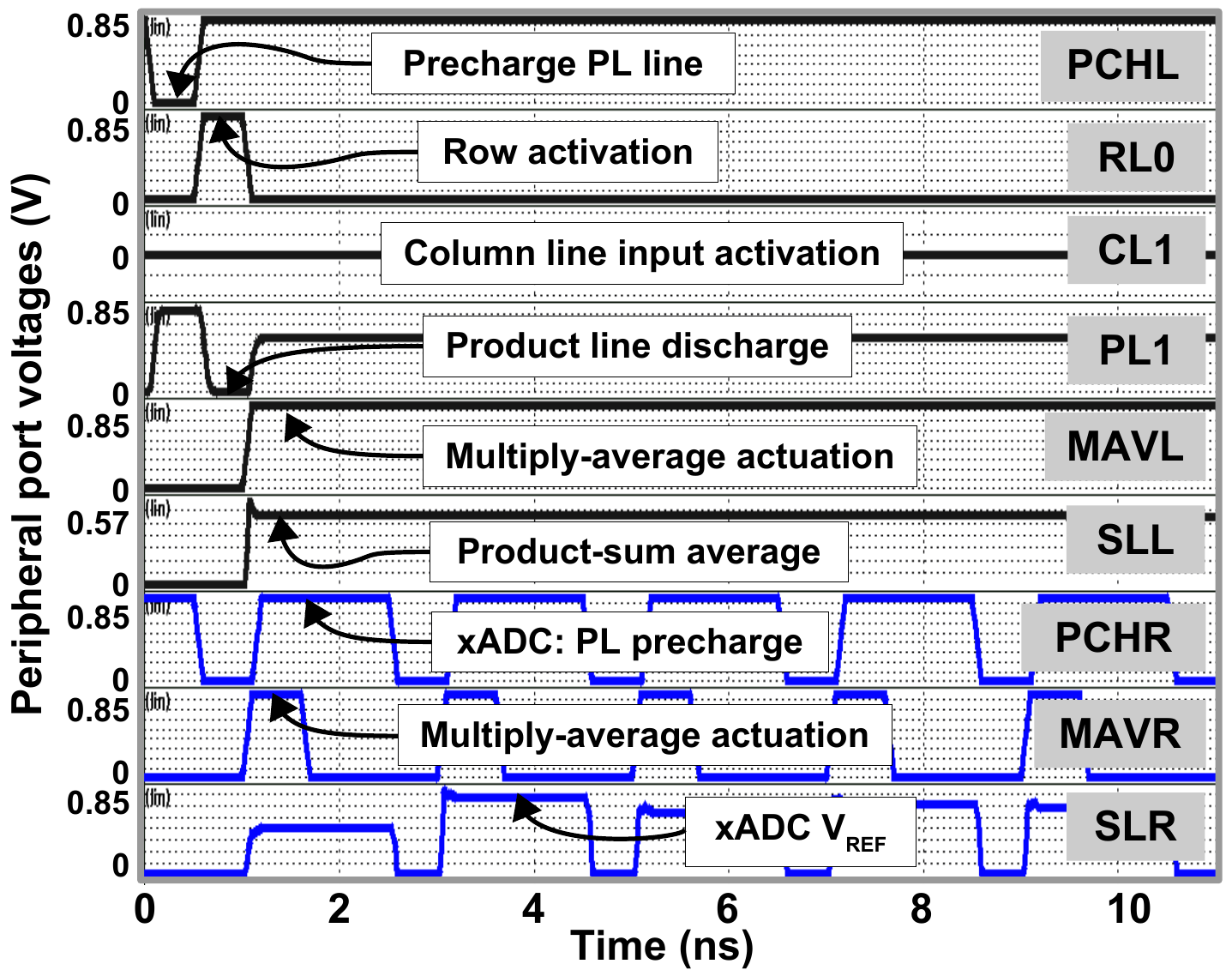}
    
    \caption{Response flow of signals in MC-CIM's bitplane-wise processing.}
    \label{fig:transient}
\end{figure}

The operation within the CIM module in Figure \ref{fig:syslevel}(c) begins with precharging PL and applying input at CL in the first half of a clock cycle. In the next half of a clock cycle, RL is activated to compute the product bit on PL port. PL discharges only when input and stored bit are both one. Figure \ref{fig:transient} shows the response and flow of various signals in 16$\times$31 SRAM-CIM macro upon input activations. The CIM macro is designed and simulated using low standby power (LSTP) 16 nm CMOS predictive technology models from \cite{sinha2012exploring}.

\begin{figure}[t]
    \centering
    \includegraphics[width=0.75\columnwidth]{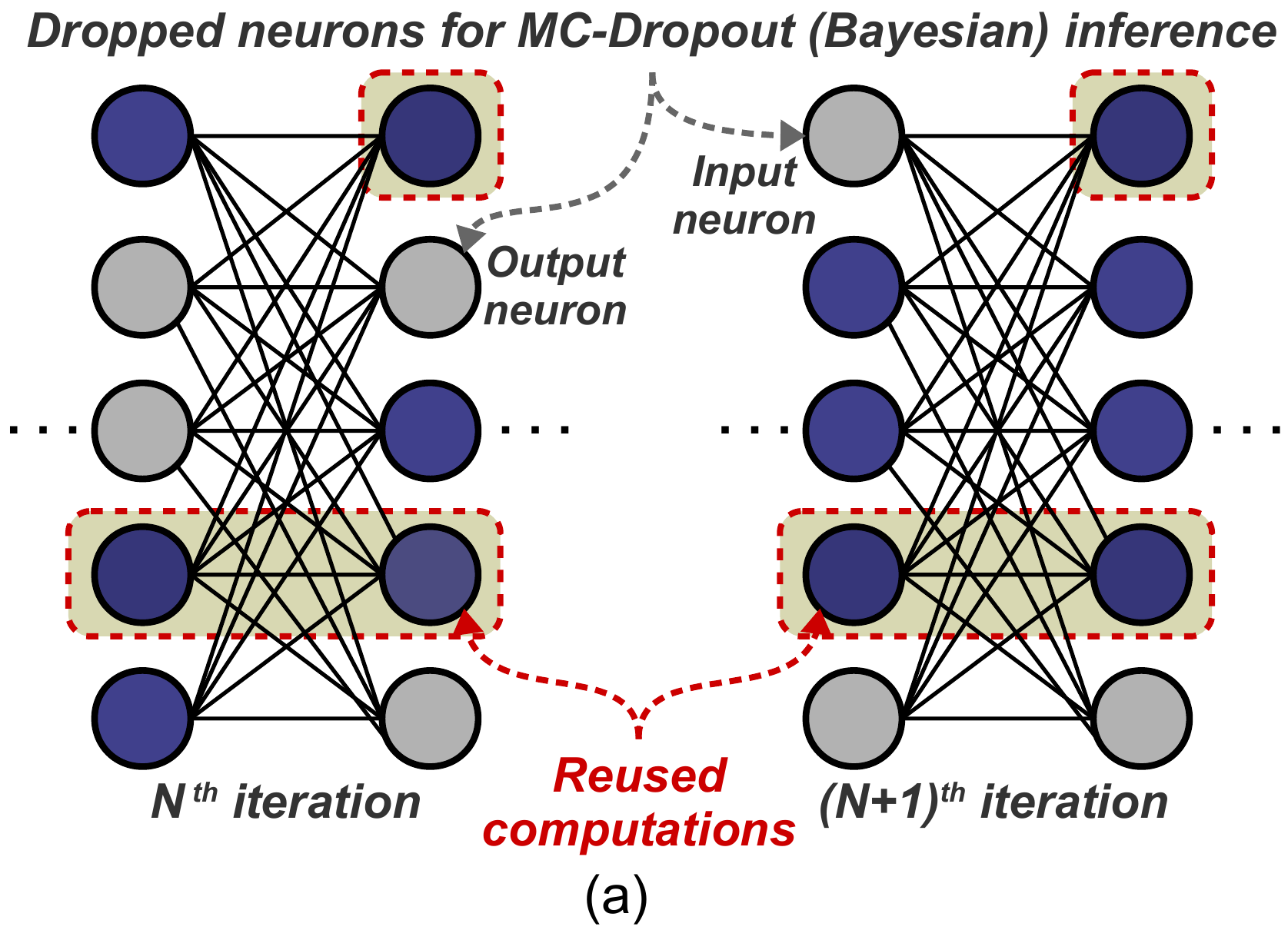}
    
    \includegraphics[width=0.825\columnwidth]{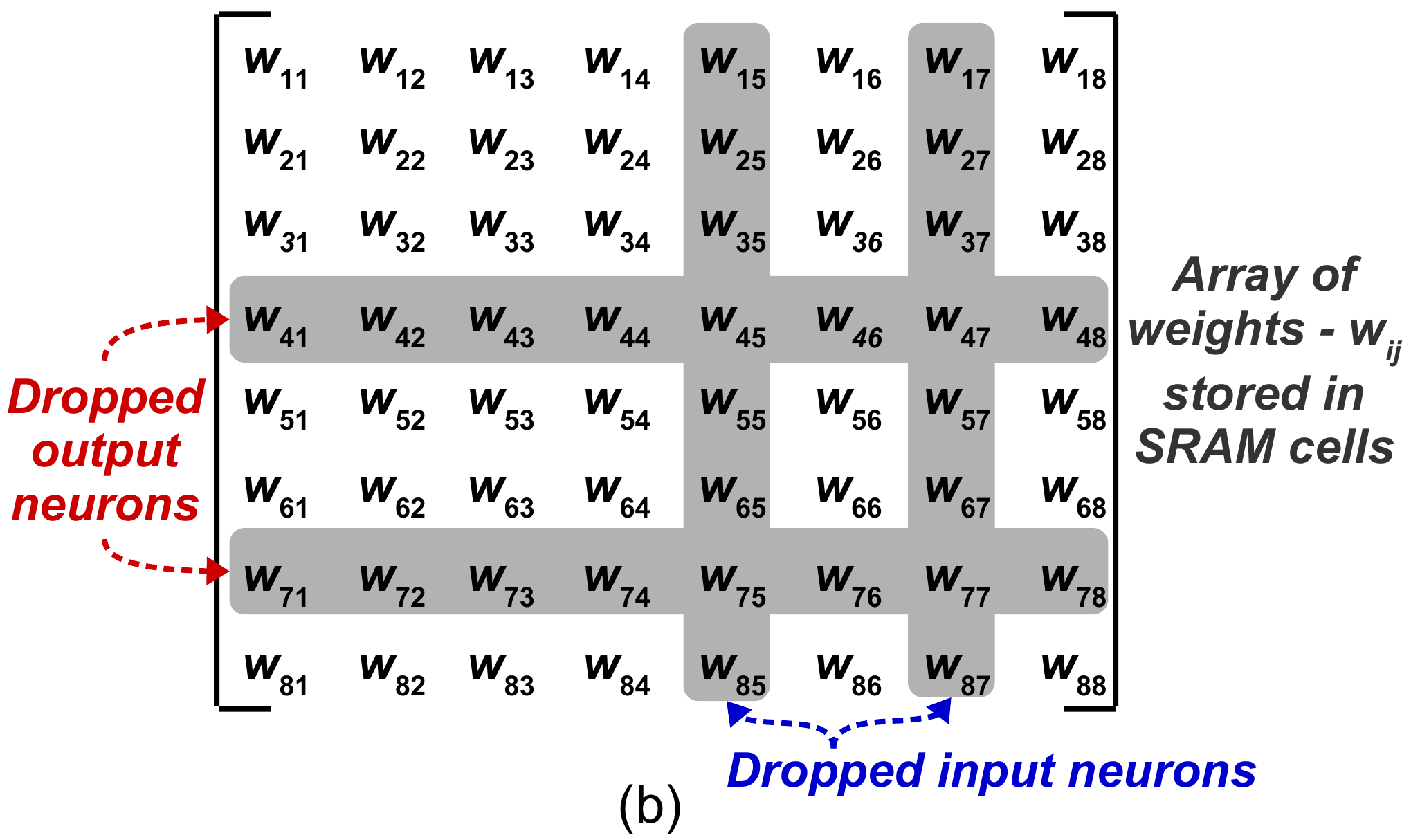}
    \caption{(a) MC-Dropout-based variational Bayesian inference for deep learning with computation-reuse for successive MC-Dropout iterations. (b) MC-Dropout mapped to CIM. Input and output neurons are dropped by deactivating corresponding rows and columns respectively in the CIM.}
    \label{fig:dropout}
\end{figure}

The output of all PL ports are averaged on the sum line (SLL) using transmission gates in the figure, determining the net multiply-average (MAV) of bitplane-wise input and weight vector. The charge-based output at SLL is passed to SRAM-immersed analog-to-digital converter (xADC), discussed in details in our prior work \cite{nasrin2021mf}. xADC operates using successive approximation register (SAR) logic and essentially exploits the parasitic bitline capacitance of a neighboring CIM array for reference voltage generation. Operating waveforms of xADC are shown in Figure \ref{fig:transient} in blue. In the consecutive clock cycles different combinations of input and weight bitplanes are processed and the corresponding product-sum bits are combined using a digital shift-ADD. 

Importantly, compared to \cite{nasrin2021mf}, we have also uniquely adapted xADC's convergence cycles by exploiting the statistics of MAV leading to a considerable improvement in its time and energy efficiency. These details are presented subsequently. 

\begin{figure*}[t]
    \centering
    \includegraphics[width=0.95\linewidth]{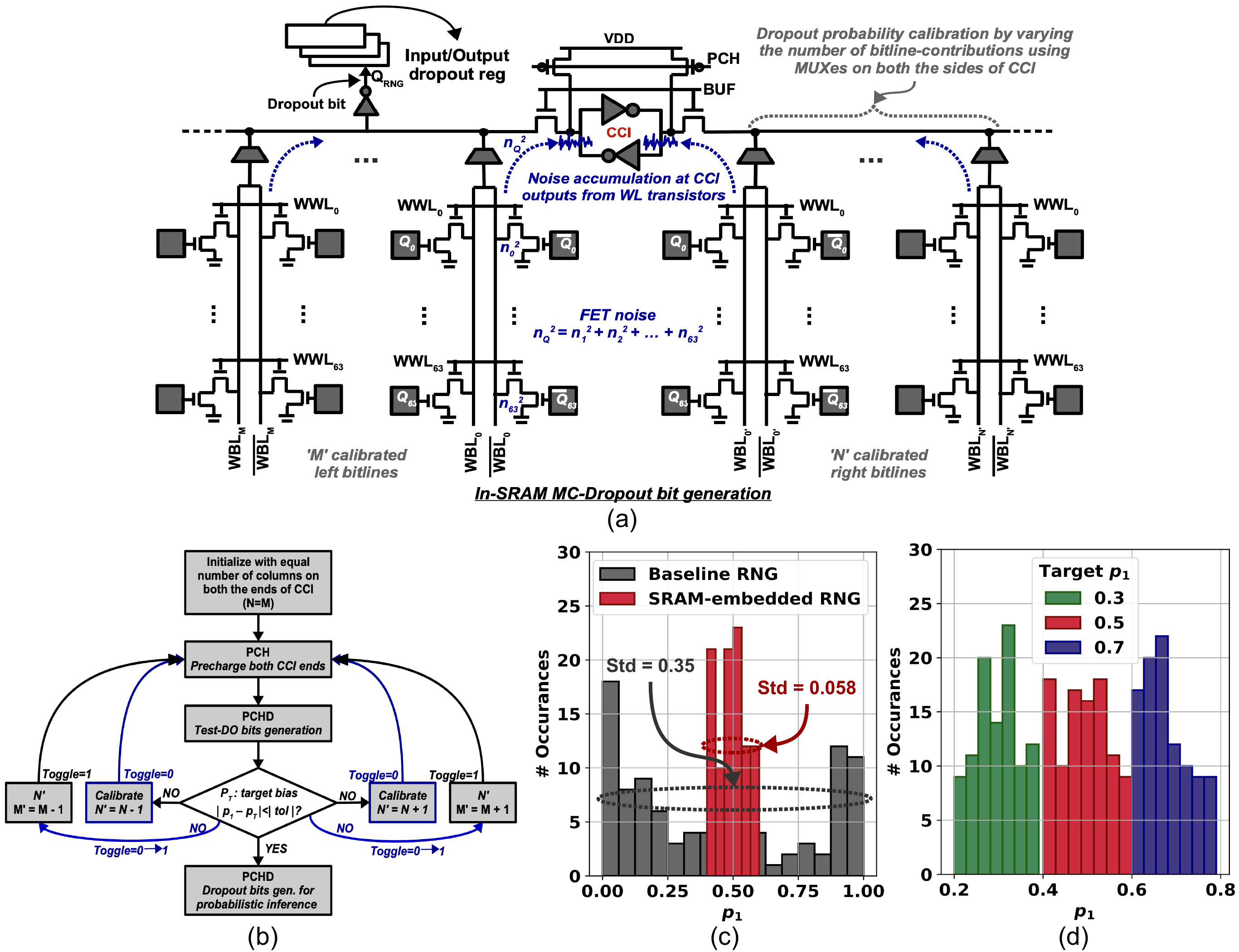}
    
    \caption{(a) SRAM-embedded random dropout bit generator. (b) Dropout probability calibration in CIM-embedded RNG. (c) Comparing baseline CCI and proposed SRAM-embedded CCI (RNGs) for random dropout bit generation probability ($p_1$) amidst transistor mismatches. (d) Tuning SRAM-embedded RNG for desired dropout probabilities (0.3, 0.5 and 0.7). Histograms correspond to 100 Monte Carlo instances.}
    \label{fig:rng}
\end{figure*}

\section{Variational Inference in Compute-in-Memory}
\subsection{MC Dropout-based Inference in Neural Networks}

Figure \ref{fig:dropout}(a) shows the high-level overview of MC-dropout-based Variational inference as presented in \cite{gal2016dropout}. In a dropout (DO) layer, input and output neurons are dropped randomly in each iteration following a Bernoulli distribution. Implementing MC-dropout layers with dropout probability of 0.5 has shown to adequately capture model uncertainties for robust inference in many applications \cite{kendall2016modelling, gal2016dropout}. Although in \cite{gal2017concrete}, authors have also developed learning schemes to extract the optimal dropout probabilities from training data. Considering the mapping of neural network weights on a CIM array, in Figure \ref{fig:dropout}(b), randomly dropping input neurons is equivalent to masking the weight operation in the corresponding column of CIM, whereas dropping output neurons is equivalent to disabling corresponding weight rows of CIM. Ensemble of several DO operations is equivalent to Monte Carlo estimates of network weights sampled from the posterior distribution of models as shown in \cite{gal2016dropout}. Predictions in such MC-Dropout-based inference procedure are made by averaging the model outputs for regression tasks and by majority voting on classification tasks. Whereas the model confidence is extracted by estimating the variance of probabilistic outputs from each iteration. 

\begin{figure*}[t]
    \centering
    \includegraphics[width=0.85\linewidth]{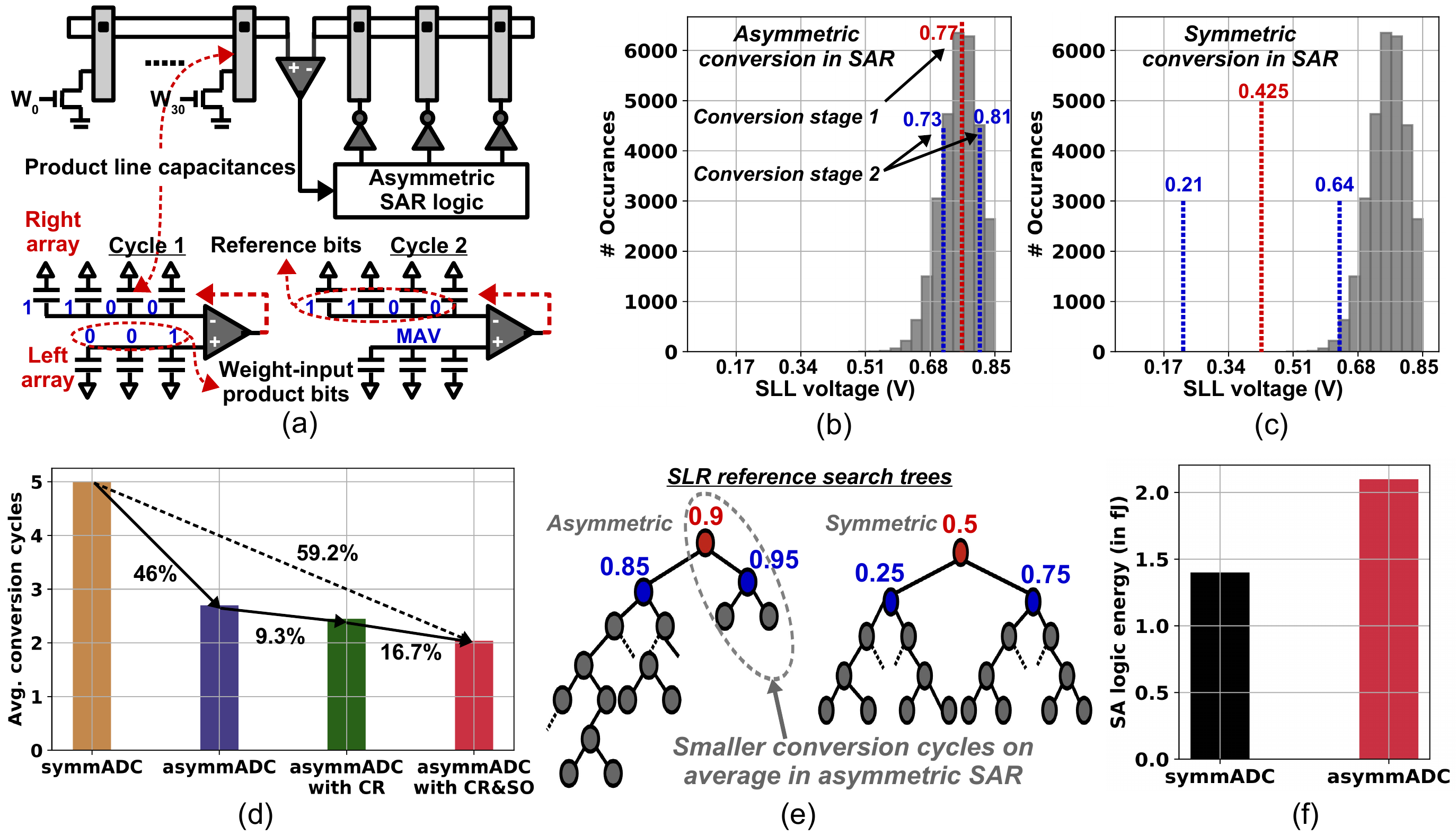}
    
    \caption{(a) SRAM-immersed analog-to-digital converter (xADC). (b-c) Product-Sum line (SLL) histograms for computation within MC-CIM with asymmetric and symmetric SAR conversioncycles. (d) Conversion cycle savings in various SAR conversion and computation modes in MC-CIM. (e) Asymmetric and symetric search for $V_{REF}$ in SAR. Red node - $1^{st}$ cycle, blue nodes - $2^{nd}$ cycle. (f) Symmetric and asymmetric SA logic energies.}
    \label{fig:adc}
\end{figure*}

\subsection{Within-SRAM Dropout Bit Generators}
In Figure \ref{fig:syslevel}(c), to support random input dropouts, inputs to CL peripherals are ANDed with a dropout bitstream. Likewise, for random output dropouts, row activations are masked by ANDing RL signals with output dropout bitstream. Therefore, inference in MC-Dropout requires an additional processing step of dropout bit generation for each applied input vector. High-speed generation of dropout bit vectors is thereby a critical overhead for CIM-based MC-Dropout. 

In prior works, random number generators (RNG) are implemented by using analog circuits to amplify noise \cite{petrie2000noise}, by exploiting uncertain phase jitters in the oscillators \cite{sunar2006provably}, by using metastability resolution in cross-coupled inverters by thermal noise \cite{mathew20122}, and by using continuous/discrete-time chaos \cite{pareschi2010implementation}. Among the prior techniques, a cross-coupled inverter (CCI) circuit is lightest-weight design for random number generation. However, the entropy of generated random bits from CCI is greatly affected by transistor mismatches. Without calibration, most of CCI instances will either show no randomness at all or a significant bias in the generated random bit stream \cite{mathew20122}. In \cite{mathew20122}, high entropy in random number generation is achieved by incorporating programmable PMOS and NMOS strengths in the CCI. Further fine-grained tuning was proposed using programmable precharge-clock delay buffers. However, CCI's calibration requires significant overhead, eclipsing its area advantages. Subsequently, we discuss how SRAM-embedded random number generation can exploit memory array's parasitics to mitigate such overheads.  

Note that each weight-input correlation cycle for our CIM-optimal inference operator ($\oplus$) lasts $2(n-1)$ clock periods for $n$-bit precision weights and inputs. Therefore, for $m$-column CIM array, a throughput of $\frac{m}{2(n-1)}$ random bits/clock is needed. Meeting this requirement, $\lceil \frac{m}{2(n-1)}\rceil$parallel CCI-based RNGs are embedded in a CIM array in our scheme, each capable to generate a dropout bit per clock period. CCI-based dropout vector generation is pipelined with CIM's weight-input correlation computations, i.e., when CIM array processes an input vector frame, memory-embedded RNGs sample dropout bits for the next frame. 

Figure \ref{fig:rng}(a) shows the proposed SRAM-embedded RNG. We exploit SRAM's write parasitics for RNG calibration. During inference, write wordlines (WWL) to a CIM macro are deactivated. Therefore, along a column, each write port injects leakage and noise current to the bitline as shown in Figure \ref{fig:rng}(a). Even though the leakage current from each port, $I_{leak,ij}$, varies under threshold voltage ($V_{TH}$) mismatches, the accumulation of leakage current from parallel ports reduces the sensitivity of net leakage current at the bitlines, i.e., $\sum_i I_{leak,ij}$ shows less sensitivity to $V_{TH}$ mismatches. Each write port also contributes noise current, $I_{noi,ij}$, to the bitline. Since the noise current from each port varies independently, the net noise current, $\sum_i I_{noi,ij}$, magnifies. We exploit such filtering of process-induced mismatches and magnification of noise sources at the bitlines for RNG's calibration. 

An equal number of SRAM columns are connected to both ends of CCI. Both bitlines (BL and $\overline{\text{BL}}$) of a column are connected to the same end to cancel out the effect of column data. Both ends of CCI are precharged using PCH signal and then let discharged using column-wise leakage currents for half a clock cycle. At the clock transition, pull-down transistors are activated using a delayed PCH (PCHD) to generate the dropout bit. For the calibration, CCI generates a fixed number of output random bits serially from where its bias can be estimated. A simple calibration scheme in Figure \ref{fig:rng}(b) then adapts the parallel columns connected to each end until CCI is able to meet the desired dropout bias within the tolerance. 

Figure \ref{fig:rng}(c) shows the histogram of CCI’s probability ($p_1$) to produce `1’ as the output. An ideal CCI-based RNG should have $p_1=0.5$. The histogram of $p_1$ over hundred instances is compared against a baseline CCI that doesn't exploit SRAM columns for calibration. For each CCI instance, $p_1$ is extracted based on 500 evaluations. CIM-embedded CCI shows a much limited variability of $p_1$; for SRAM-embedded CCI, $\sigma(p_1)=0.058$, whereas for baseline CCI, $\sigma(p_1)=0.35$. In Figure \ref{fig:rng}(d), we also show SRAM-embedded RNG calibrated for $p_1$ = 0.3 and 0.7 bias targets, meeting similarly tight margins. 

The operation of CCI-based dropout generation can be further improved using fine-grained calibration (such as \cite{mathew20122}) along with the coarse-grained calibration presented here. However, our later discussion in Sec. VI will show an adequate tolerance to RNG's bias perturbation in MC-Dropout which has motivated our lightweight scheme above.

\subsection{Exploiting MAV Statistics for ADC's Time Efficiency}
The probabilistic activation of inputs in MC-Dropout can also be exploited to adapt the digitization of multiply average voltage (MAV) generated at the sumline (SLL). SLL voltage in MC-CIM follows $V_{DD}-\frac{V_{DD}}{n}\sum x_i \cdot w_i$ where $x_i$ and $w_i$ are input and weight vector bits processed in column $i$, $V_{DD}$ is the supply voltage, same as the column precharge voltage, and $n$ is the number of columns in CIM array. At input dropout probability $p_1 = 0.5$, about half of the input bits are deactivated. This induces an asymmetry in MAV's voltage distribution skewed towards $V_{DD}$. We exploit this statistics of MAV to improve the time efficiency of digitization.  

In our prior work \cite{nasrin2021mf}, we presented successive approximation (SA)-based memory-immersed data converter (xADC). In Figure \ref{fig:adc}(a), bitline capacitance of a neighboring CIM array is exploited in xADC to realize the capacitive DAC for SA, thereby, averting a dedicated DAC and corresponding overhead. While xADC in \cite{nasrin2021mf} followed typical binary search of a conventional data converter, here, we discuss an \textit{asymmetric successive approximation}. The histogram in Figures \ref{fig:adc}(b-c) depict skew of MAV distribution that is asymmetrically situated closer to V$_{DD}$. We can thus minimize the digitization cycles for MAV using asymmetric approximation. For this, reference levels at each cycle are selected based on the MAV statistics such that they iso-partition the distribution segment being approximated by the conversion cycle. For example, in the first cycle, the first reference point $R_{0}$ is $\sim$mean(MAV), instead of half of V$_{max}$ where V$_{max}$ is the maximum voltage generated at sumline (SLL). Likewise, in the next iteration, reference levels $R_{00}$ and $R_{01}$ are generated to iso-partition MAV distribution falling between $[0, R_{0}]$ and $[R_{0},V_{max}]$, respectively. Since asymmetric SA results in unbalanced search of references in Figure \ref{fig:adc}(e), very few cases requires more SA cycles than in conventional SA ADC, and for the majority of inputs, the total searches are much less.  

Figure \ref{fig:adc}(d) shows the advantage of asymmetric SA over the conventional one. For a 5-bit conversion of MAV, asymmetric SA requires on average $\sim$2.7 cycles, 46\% less than the conventional. In the next section, we will discuss adaptations of MC-CIM for compute reuse (CR) and sample ordering (SO) flows which further increase input sparsity, and proportionally lead to even more benefits of asymmetric SA. In Figure \ref{fig:adc}(d), asymmetric SA with CR and SO, only requires two conversion cycles. The asymmetric SA, however, comes at the cost of higher logic complexity. In Figure \ref{fig:adc}(f), a finite state machine (FSM)-based asymmetric SA logic incurs 2.1fJ/operation on average whereas the typical SA logic incurs 1.4fJ. The energy of FSM-based SA logic was extracted using register-transfer level (RTL) synthesis and extraction using Cadence RC compiler. Nonetheless, since energy overheads of ADC are dominated by analog operations, comparator and DAC precharge, asymmetric SA is more energy efficient overall.

\section{Data-Flow Optimization in MC-CIM}
\subsection{Compute Reuse in MC-Dropout Inference}
Two successive MC-Dropout iterations in Figure \ref{fig:dropout}(a) can share common input/output neurons. Therefore, the workload in MC-Dropout can be reduced significantly by iteratively computing the product-sum in each iteration as $P_{i} = P_{i-1} + W\times I^A_{i} - W\times I^D_{i}$. Here, $I^A_{i}$ indicates the input neurons that are active in the current iteration but were dropped in the previous one. $I^D_{i}$ denotes the input neurons that were active in the previous iteration but are dropped currently. This way, we perform only the newer computations and reuse the overlapping ones.

\begin{figure}[t]
    \centering
    \includegraphics[width=\columnwidth]{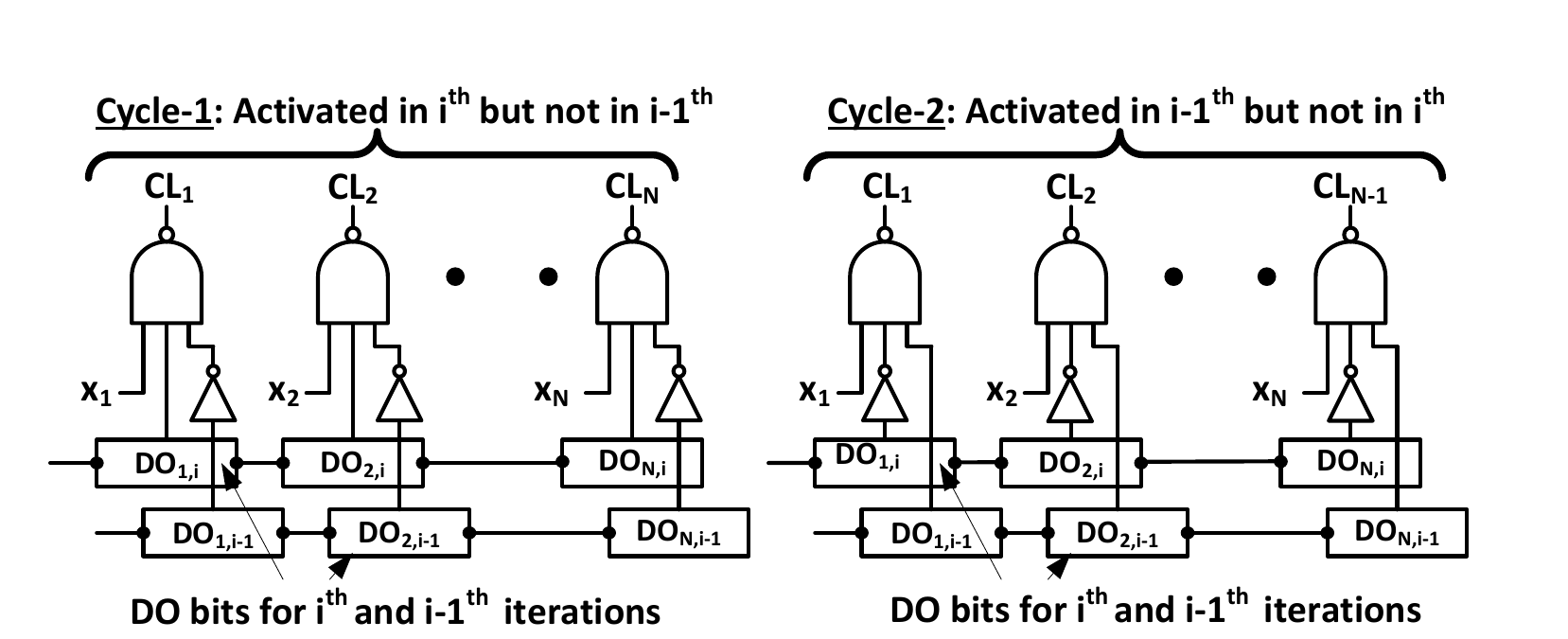}
    \caption{Logic operations to implement compute reuse.}
    \label{fig:crimplement}
\end{figure}

Figure \ref{fig:tsp}(b) shows the advantages of such compute reuse. Considering an example processing of fully-connected input and output layers with 10 neurons in each layer, the figure compares the necessary multiply-accumulate (MAC) operations for typical and compute reuse-based execution. For a MC-Dropout inference that considers hundred dropout samples for prediction, compute reuse-based execution requires only $\sim$52\% MAC operation compared to a typical flow. Figure \ref{fig:crimplement} shows the implementation of compute reuse. At each iteration, computations are performed in two cycles. In the first, cycle-1, only those activations that are present in i$^{th}$ iteration but not in i-1$^{th}$ are processed. While in second, cycle-2, activations that are present in i-1$^{th}$ iteration but not in i$^{th}$ are processed. The selection of non-overlapping activations can be made by retaining dropout bits for the previous iteration and using simple logic operations as shown in the figure.   

\subsection{Optimally Ordering Dropout Steps}
Moreover, we can make the above compute reuse even more effective by optimally ordering the dropout samples of MC-Dropout inference. The ordering has to be such that the successive iterations have the maximum overlap of active/inactive neuron set so that the cumulative workload minimizes while traversing through the entire dropout sample set.

\begin{figure}[t]
    \centering
    \includegraphics[width=\columnwidth]{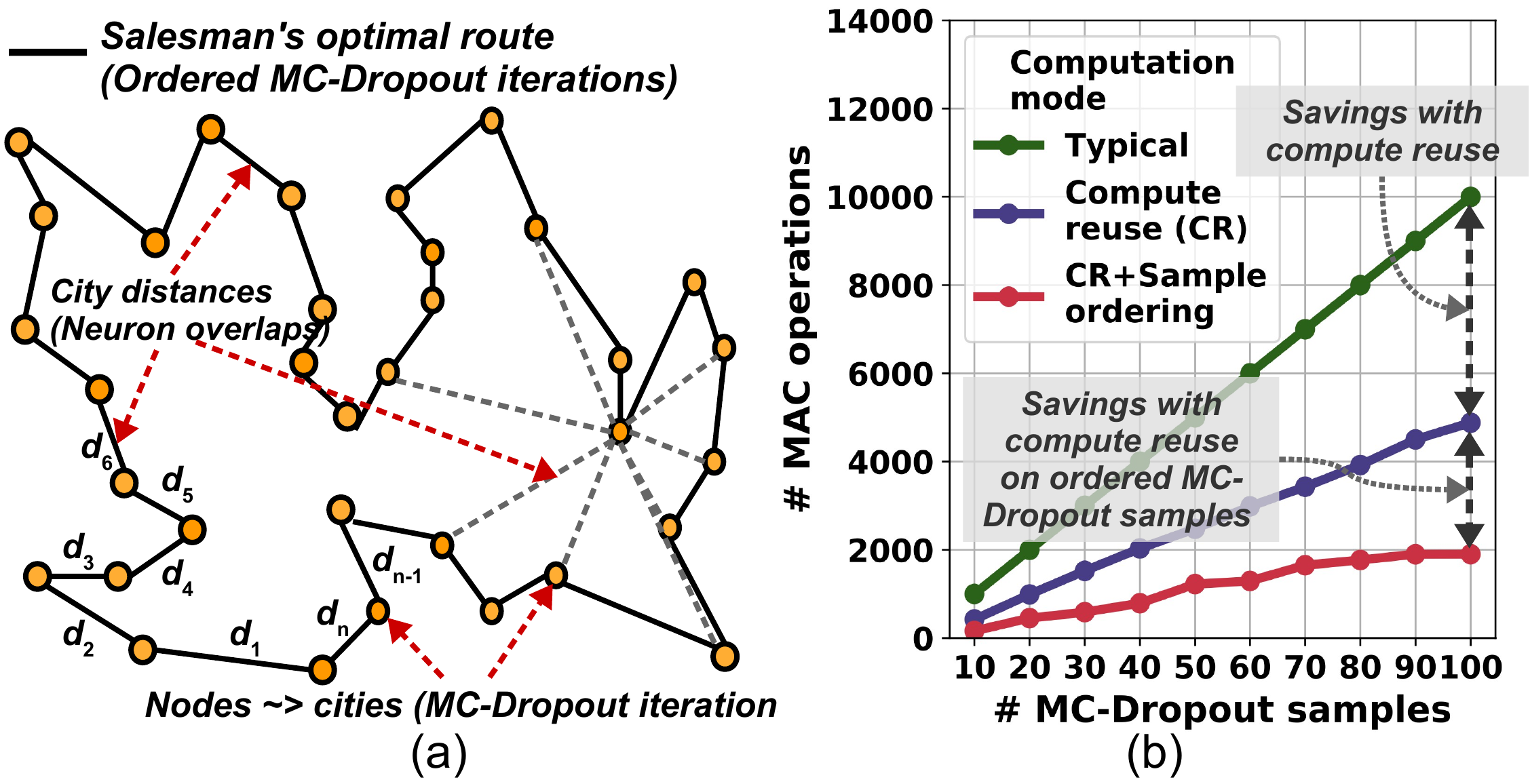}
    \caption{(a) Travelling salesman problem (TSP) to identify optimal (minimum distance travelled) path covering all cities. (b) Computation savings in MC-Dropout based BI with typical inference flow, compute reuse in successive iterations and compute reuse with TSP-based optimal sample ordering.}
    \label{fig:tsp}
    \vspace{-1em}
\end{figure}

A combinatorial optimization problem that is equivalent to our objective here is the traveling salesman's problem (TSP). TSP is a well-studied problem whose objective is to find an optimal traversal path by a salesman to visit `n' cities where each city is visited only once as shown in Figure \ref{fig:tsp}(a). In an analogy to TSP, iterations of MC-Dropout represent cities. For two samples `i' and `j', $I^A_{ij} + I^D_{ij}$ represents their distance where $I^A_{ij}$ indicates the input neurons that are active in the `i' iteration but were dropped in `j.' $I^D_{ij}$ denotes the input neurons that were active in the `j' iteration but are dropped in `i.' Even though, TSP is an NP-hard problem, several efficient optimization procedures exist for the problem \cite{halim2019combinatorial}. Using these combinatorial optimization procedures, an optimal dropout sequence can be computed in advance and embedded in the inference flow. Figure \ref{fig:tsp}(b) compares computation savings in typical, compute reuse and compute reuse with optimal sample ordering. Compared to typical inference flow, the computation savings for a hundred MC-Dropout samples using compute reuse with TSP-based optimal sample ordering is even better, $\sim$80\% in the figure.

Also note that the above strategy of precomputing and optimally ordering dropout bits obviates SRAM-embedded TRNGs. However, instead, additional storage of ordered dropout schedules becomes necessary. These dropout schedules can be stored in an additional SRAM interfaced with CIM to sequentially readout during MC-Dropout. Although the storage of dropout schedules is an important overhead of the sample-ordering method, \cite{kendall2016modelling} shows that sufficient output statistics can be extracted from few ($\sim$10-30) dropout iterations making pre-computation of samples and their ordering an attractive alternative for many applications.

\begin{figure}[t]
    \centering
    \includegraphics[width=0.6\columnwidth]{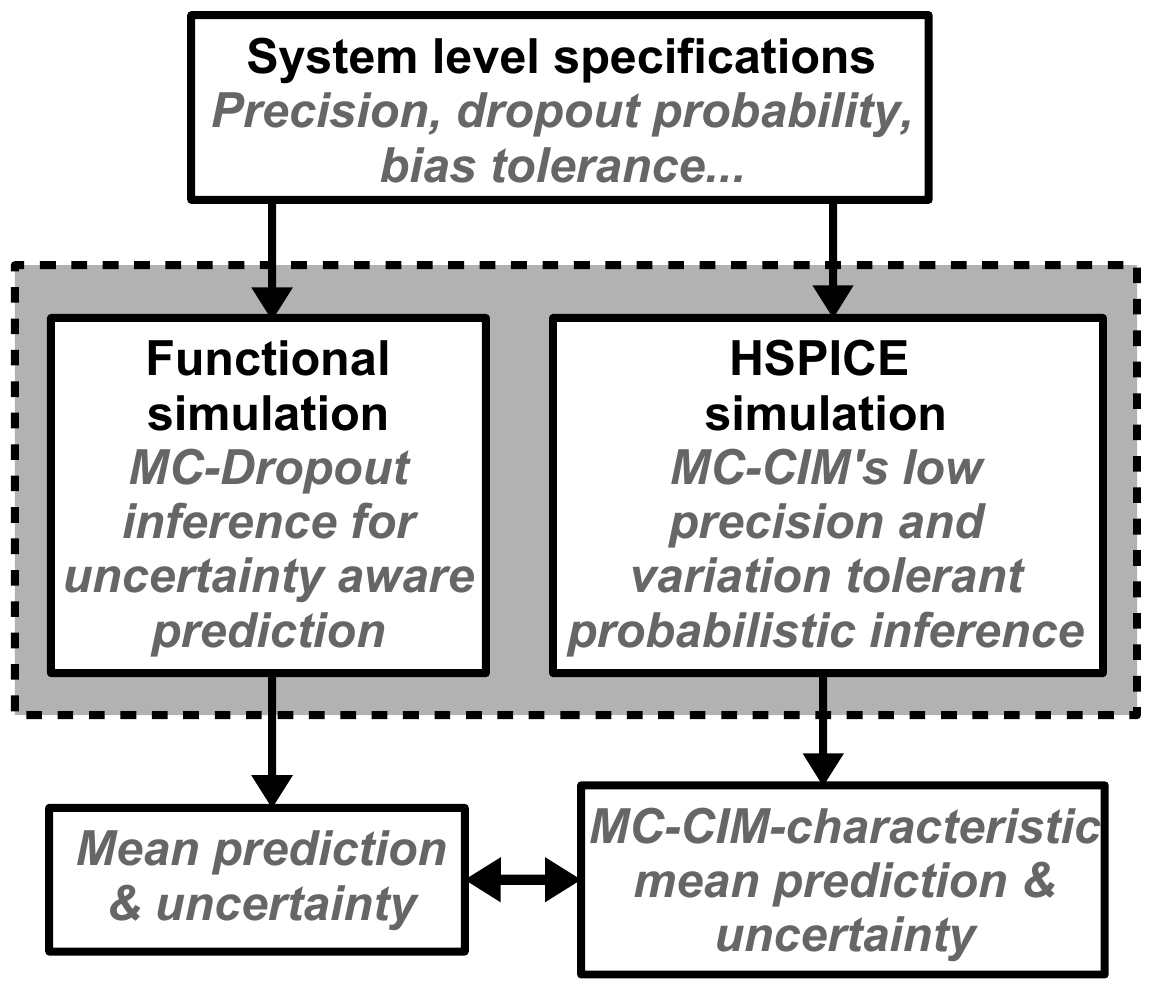}
    
    \caption{Simulation methodology projecting MC-CIM macro's characteristics to Bayesian deep neural network (DNN) inference.}
    \label{fig:sim_method}
\end{figure}

\section{Power-Performance Characterization of CIM-based MC-Dropout Macro}
\subsection{ Methodology to Project Macro-Characteristics to System}

\begin{figure}[hbt!]
    \centering
    \includegraphics[width=0.8\columnwidth]{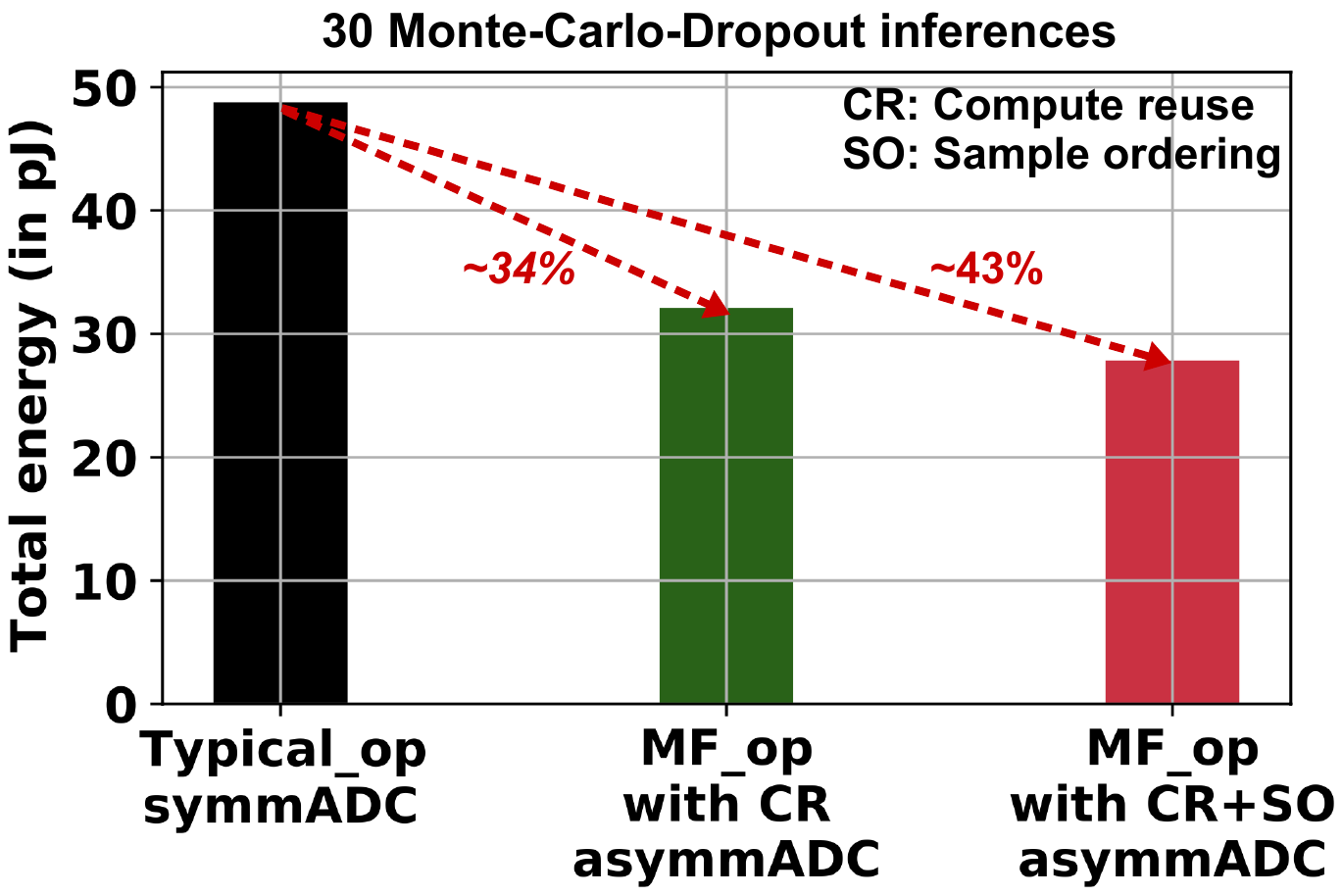}
    \caption{Energy comparison at various operating modes of MC-CIM for MC-Dropout inference with 30 dropout iterations per input and 6-bit precision.}
    \label{fig:energy_comp}
\end{figure}

\begin{figure}[hbt!]
    \centering
    \includegraphics[width=\columnwidth]{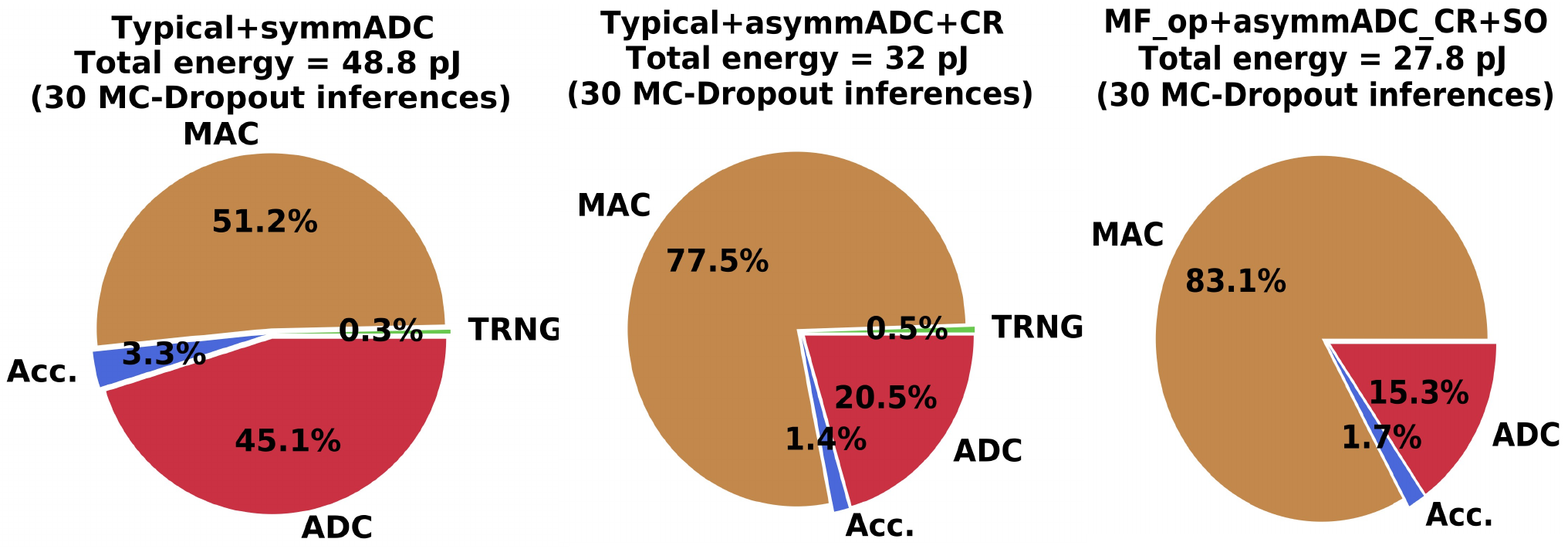}
    \caption{Energy breakdown comparisons in various operating modes of $16\times31$ MC-CIM macro for MC-Dropout inference at 6-bit precision.}
    \label{fig:pie_energy}
\end{figure}

\begin{figure}[hbt!]
    \centering
    \includegraphics[width=\columnwidth]{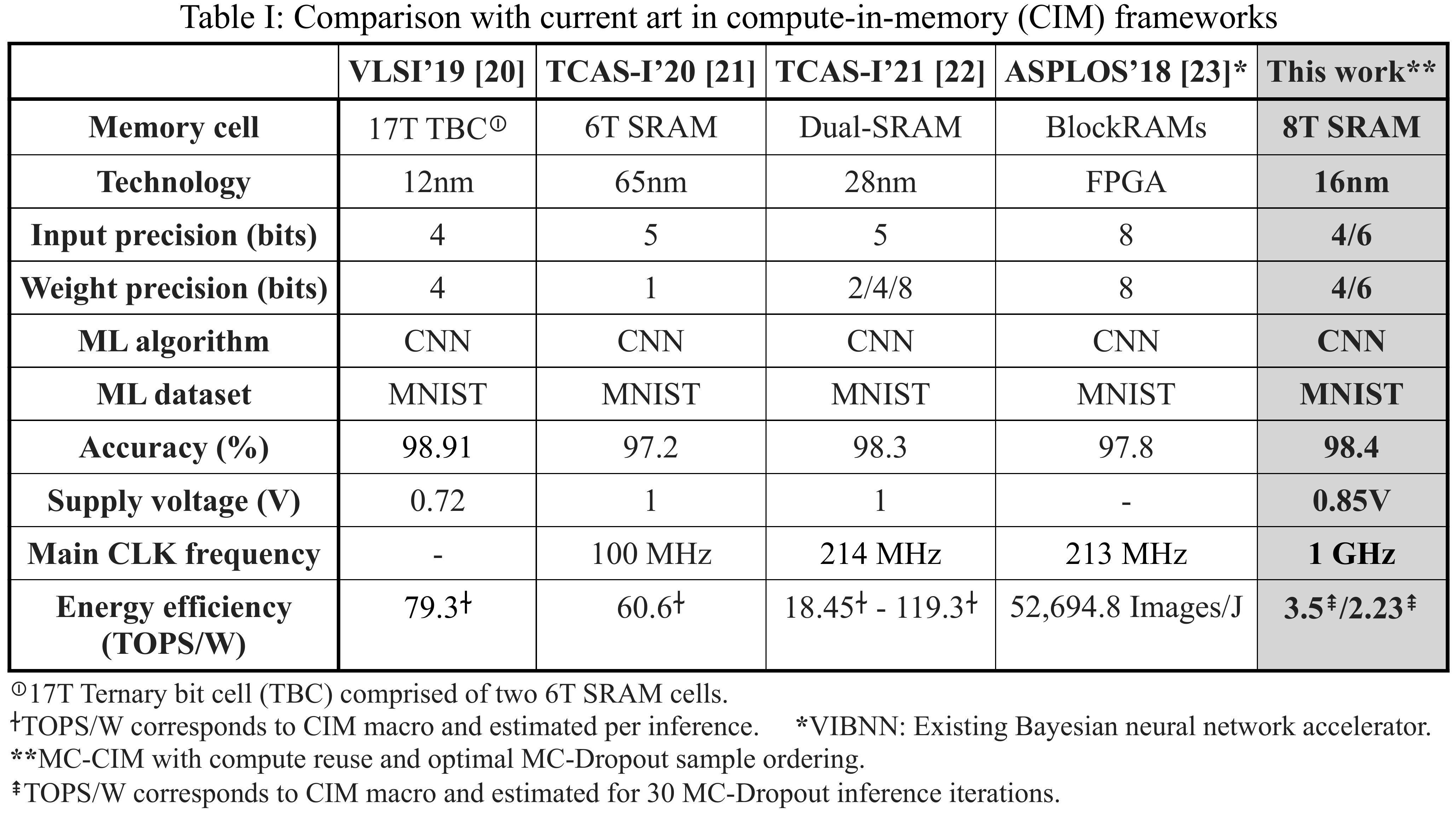}
    \label{fig:table}
\vspace{-1em}
\end{figure}

\begin{figure}[t]
    \centering
    \includegraphics[width=0.9\columnwidth]{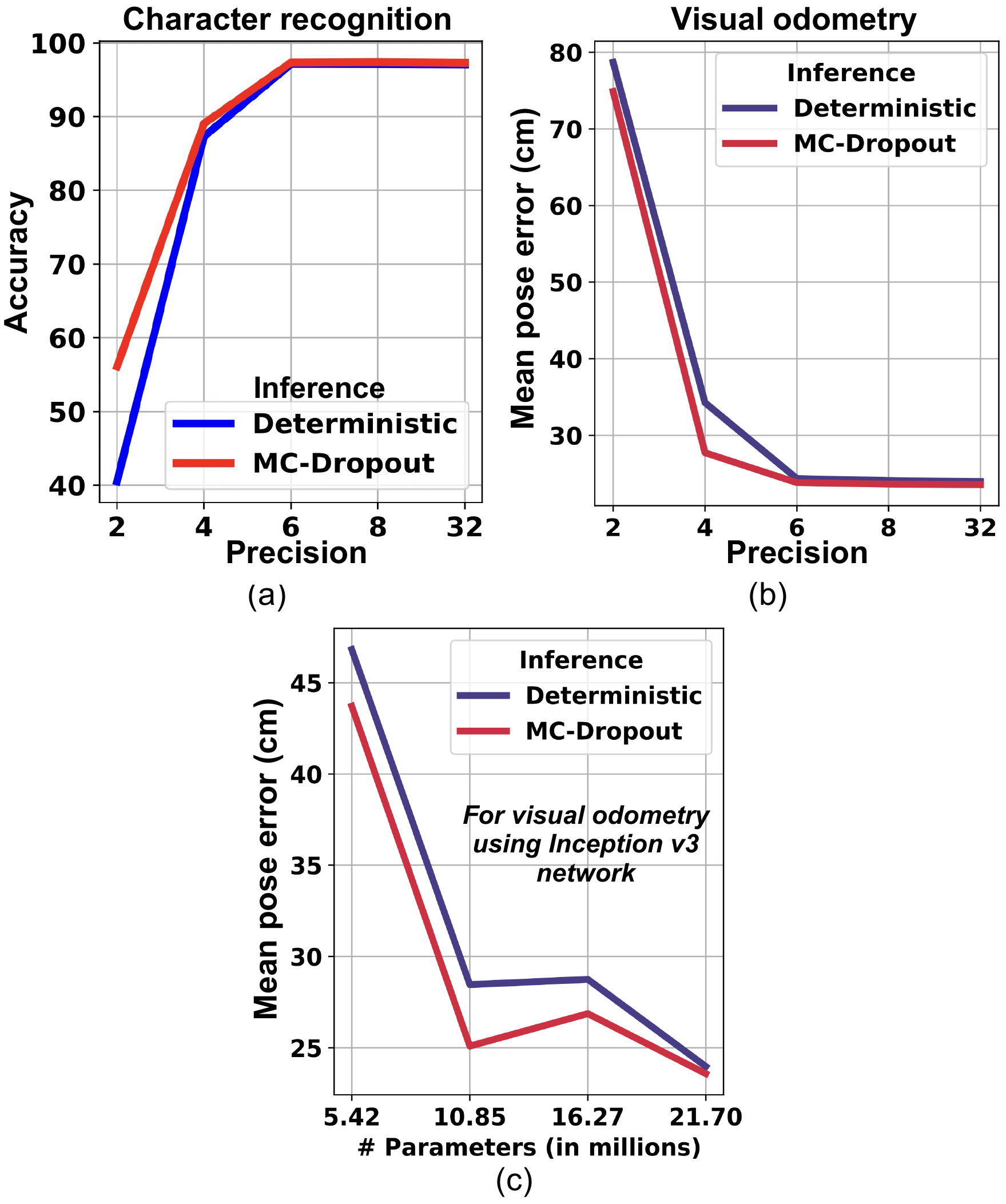}
    \caption{Precision-Accuracy comparison between deterministic and MC-Dropout inference for (a) character recognition, and (b) visual odometry (VO). (c) Impact of parameter-reduction on accuracy for VO.}
    \label{fig:accplots}
\end{figure}

In Figure \ref{fig:sim_method}, we present our methodology to project CIM's macro-level characteristics to system-level benchmarks. First, a full precision MC-Dropout-based DNN model is downgraded to CIM's lower input and weight precision. Next, process variability-induced non-idealities in RNG-generated dropout bits are accounted by adding statistical perturbation to dropout probabilities as shown in the Figure \ref{fig:mnist}(c). The perturbation statistics are extracted from Macro-level SPICE simulations at varying power-constraints. For example, as fewer SRAM columns are used for RNG calibration (using the circuit in Figure \ref{fig:rng}(a)), a higher deviation in dropout probabilities is observed. Conversely, operating a RNG with fewer SRAM columns results in lower power operation since the net capacitance at CCI ends decreases. Such deviations in dropout probabilities are fitted with a Beta distribution, as shown in Figure \ref{fig:mnist}(c), and distribution parameters are used in the perturbation layers to study the system-level power scaling.

\subsection{Energy characterization of MC-CIM}
We characterize MC-CIM's total energy for MC-Dropout-based probabilistic inference by considering the energy consumed for random number generation, product-sum computation, analog-to-digital conversion and accumulation of product-sums. In Figure \ref{fig:energy_comp} we compare the energy consumption by MC-CIM for MC-Dropout inference with 30 dropout iterations per input. Compared to conventional configuration of typical operator and typical ADC, MC-CIM with multiplication-free (MF) operator, asymmetric successive approximation and MAC with compute reuse saves $\sim$34\% energy consuming 32 pJ. Energies for different configurations of MC-CIM are shown in the bar plot in Figure \ref{fig:energy_comp}. With compute reuse as well as additionally incorporating optimal sample ordering, energy consumption reduces even more to 27.8 pJ. Figure \ref{fig:pie_energy} shows the energy breakdown for various peripheral operations in MC-CIM at three different configurations as shown in the figure. An interesting note from the figure is that while in majority of CIM approaches the net energy is dominated by ADC, such as also for the typical processing in the left most pie-chart, with compute reuse and MAV statistics-aware time-efficient ADC, the proportion of ADC's contribution to total energy reduces in compute reuse case to $<$21\% and compute reuse with sample ordering case to to $<$16\%.

In Table I, we compare MC-CIM with current arts in CIM frameworks \cite{okumura2019ternary, liu2020ns, lee2021charge, cai2018vibnn}. MC-CIM operates at 1 GHz clock, 0.85V and 6-bit input/weight precision to maintain state-of-the-art accuracy on various inference benchmarks. The energy efficiency of MC-CIM in its with MF operator, compute reuse, SRAM-embedded dropout-bit generation is $\sim$3.04 TOPS/W (at 4-bit precision) and $\sim$2 TOPS/W (at 6 bits) for 30 MC-Dropout network iterations whereas that for the most optimal configuration incorporating TSP-based optimal MC-Dropout sample ordering (and thus offline dropout-bits sampling) is $\sim$3.5 TOPS/W (at 4 bits) and $\sim$2.23 TOPS/W (at 6 bits). Compared to the other benchmarks in the table, although TOPS/W in our design is lower, note that our design considers Bayesian inference where each prediction is made by averaging predictions from thirty iterations and can extract both the prediction and the confidence on prediction, unlike other benchmarks in the table which only consider a classical inference of deep learning models. In Table I, we also compare MC-CIM with existing Bayesian neural network (BNN) accelerator, VIBNN \cite{cai2018vibnn}, which is FPGA-based BNN implementation with the energy efficiency of 52,694.8 Images/J predicting MNIST images with 97.81\% accuracy at 8-bit precision. 

\subsection{Synergy between probabilistic inference and MC-CIM}
Bayesian inference methods such as MC-Dropout are critical for the next generation risk-aware applications. Our previous discussion shows a significant synergy between Bayesian inference and MC-CIM, suggesting compute-in-memory as a promising pathway for robust intelligence in edge devices. Typical Bayesian inference methods operate by sampling. By coalescing sampling and inference on each sample within the same physical structure, compute-in-memory reduces data movements and energy cost for probabilistic iterations. In Figure \ref{fig:accplots}, we show the prediction accuracy curves for character recognition and visual odometry at various input/weight precision of MC-CIM. Both applications and their networks are discussed in more details later. Notably, compared to deterministic deep learning model, MC-CIM is more precision scalable, showing better accuracy at 4-bit precision. Even more, In Figure \ref{fig:accplots}(c), when using a thinner network with fewer parameters for VO, we find that Bayesian inference maintains a better accuracy than deterministic inference. Therefore, while high area demand is an important challenge for compute-in-memory methods compared to typical digital accelerators which utilize more density efficient memory units, by allowing better precision scalability, Bayesian inference also reduces the storage cost for CIM implementation of the method, thereby it is synergistic with CIM.    

\begin{figure*}[t]
    \centering
    \includegraphics[width=0.85\linewidth]{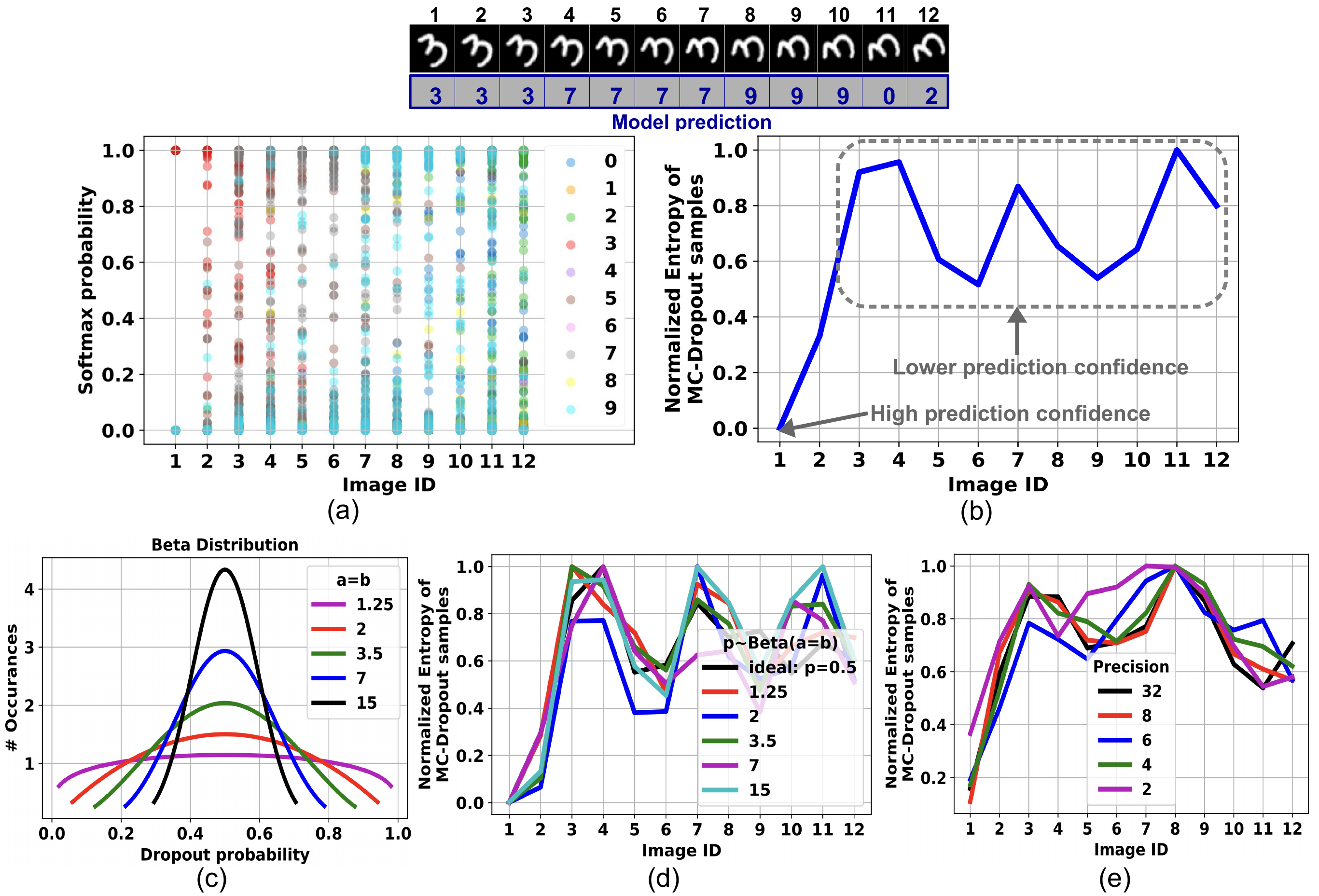}
        \caption{(a) Distribution of output classes for multiple rotations of digit 3. (b) Normalized entropy of all MC-Dropout (Bayesian) samples summed for 12 image rotations quantifying uncertainty in predictions. (c) Beta distributions for various 'a'. (d) Entropy with various dropout probability perturbations. (e) Entropy at various input/weight precisions.}
    \label{fig:mnist}
\end{figure*}

\begin{figure*}[]
    \centering
    \includegraphics[width=0.8\linewidth]{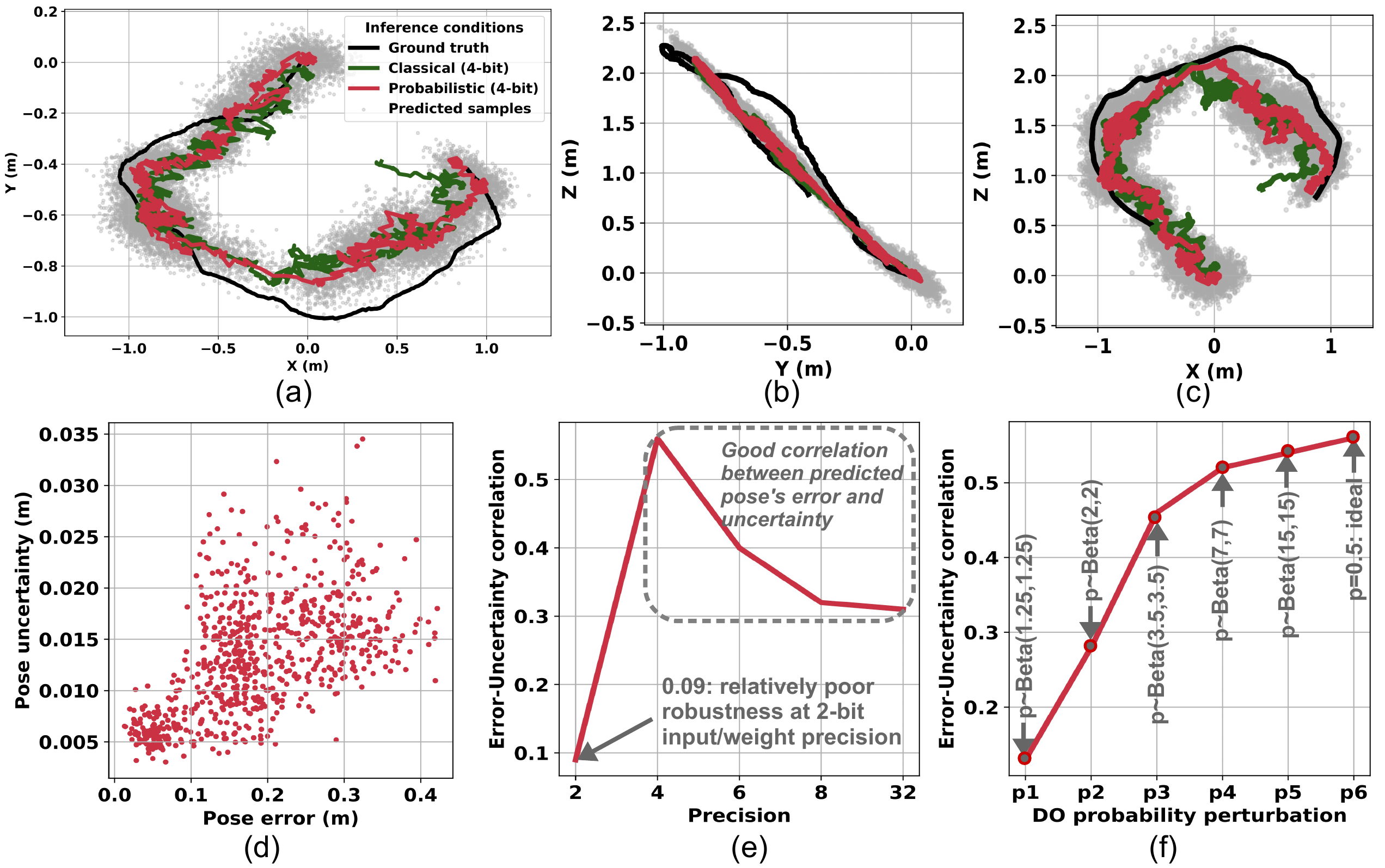}
    \caption{MC-Dropout samples based indoor (RGB-D dataset) localization and trajectory of drone in (a) X-Y, (b) Y-Z and (c) X-Z position coordinates, and comparison with deterministic network configurations at various inference conditions (a deterministic inference with 32-bit and 4-bit precision input and weights, and MC-Dropout inference at 4-bit precision inputs and weights and 30 Bayesian samples per image frame). (d)  Confidence bounds of MC-Dropout based pose samples around the mean pose prediction. (e) Correlation between the uncertainty (variance) and the error in the pose estimates. (f) Implication of PVT variation induced non-idealities in dropout probability of error-variance correlation.}
    \label{fig:poseplots}
\end{figure*}

\section{Confidence-Aware Inference under Uncertainties with MC-CIM}

We next study \textit{confidence-aware inference} under uncertainties with MC-CIM. Two benchmarking applications are considered. \textit{First}, for character recognition using MNIST \cite{lecun1998gradient}, we study MC-CIM's capacity to express predictive uncertainties when the input images are disoriented. Since the original network was trained only on the standard training images in MNIST dataset, a higher order of disorientation in input images should lead to MC-CIM expressing lower confidence (higher uncertainty) in its predictions. \textit{Secondly}, for visual odometry (VO) in autonomous drones, we evaluate a correlation factor between the error from the ground truth and MC-CIM's estimate of predictive uncertainty. With a strong correlation between the error and MC-CIM's predictive uncertainty, a higher variance in probabilistic estimates will indicate lower confidence in predictions. Thereby, downstream decision and control units can account for MC-CIM's prediction confidence for planning risk-aware drone control.

\subsection{Predictive Uncertainties under Character Disorientation}
In Figure \ref{fig:mnist}, we evaluate MC-CIM's predictive uncertainty in digit recognition by forwarding through twelve different rotation configurations of handwritten digit `3.' For each input, MC-CIM is operated 30 times, each time employing a random dropout of neurons. Our previous simulation methodology, integrating macro-level SPICE simulations with network-level functional simulation in Sec. V, is followed. In Figure \ref{fig:mnist}(a), a scatter plot of output classes over twelve different rotations is prepared where with increasing image-ID, the input image is disoriented to a higher degree. As evident in the figure, with the original image (Image-ID: 1), all thirty iterations evaluate the correct output class, and thereby, MC-CIM is highly confident in its prediction. However, with increasing Image-ID, as the input image is disoriented to a higher degree, MC-CIM's predictions are dispersed over many output classes which indicates a lower prediction confidence.    

The uncertainty in MC-CIM's predictions can be quantitatively extracted by evaluating the cross-entropy, i.e., $-\sum p_i \times \text{log}(p_i)$ where $p_i$ is the output probability for the $i^{th}$ digit class, as shown in Figure \ref{fig:mnist}(b). $p_i$ is extracted by dividing the occurrence of $i^{th}$ class in the ensemble from the total number of dropout iterations. Cross-entropy will be small when one of the output class is dominant, i.e., when MC-CIM is quite confident in its prediction. Cross-entropy will be high when none of the output class is dominant, indicating lower prediction confidence. 

Compared to the functional evaluations of MC-Dropout for such confidence-aware predictions, two main non-idealities that affect MC-CIM are: (i) low precision of weights and inputs, and (ii) process-induced bias in embedded RNGs. In Figure \ref{fig:mnist}(c-e), we systematically study these two factors. First, to study the system-level impact of RNG bias, instead of choosing the ideal dropout bias ($p$=0.5), we sample it from a symmetric Beta distribution, $p \sim\mathcal{B}(a,a)$. With decreasing $a$, the variance of Beta distribution increases (Figure \ref{fig:mnist}(c)), therefore, the sampled bias $p$ deviates much more from the ideal, emulating non-idealities of a circuit implementation of RNG. Incorporating such random bias perturbation, in Figure \ref{fig:mnist}(d), we again extract the cross-entropy characteristics from MC-CIM similar to Figure \ref{fig:mnist}(b). Despite selecting a very small $a$, i.e., considering a very high degree of RNG non-ideality, the cross-entropy curves do not deviate much from the ideal, thereby indicating a higher degree of tolerance to RNG non-idealities in MC-CIM. In Figure \ref{fig:rng}, we have exploited this attribute to simplify RNG design by considering only CIM-embedded coarse-grained calibration. In Figure \ref{fig:mnist}(e), we show the affect of low input and weight precision in MC-CIM on the probabilistic inference. Except at 2-bit precision where the entropy is high ($\sim$0.4) even for the Image-ID: 1, not much deviation is observed in the entropy curves for 4-bit precision onward and thus indicating MC-CIM's tolerance to low-precision for probabilistic inference.

\subsection{Confidence-Aware Self-Localization of Drones}
Visual odometry (VO) is a localization framework where the position and orientation of a moving camera, such as mounted on a drone, is estimated based on input image sequences. Using VO, an autonomous drone can localize itself from visual inputs. In recent years, deep learning methods have become popular for VO. A DNN-based regression network, PoseNet, was shown in \cite{kendall2015posenet} showing high accuracy for VO with a lightweight and easily trainable network. In \cite{shukla2021ultralow}, we presented floating-gate transistor based CIM framework for localization using classical particle filtering. Here, we test PoseNet under variational execution and with MC-CIM's non-idealities. 

For our study, the network was trained with scenes 1, 2 and 3 of RGB-D scenes v2 dataset \cite{lai2014unsupervised} and tested with scene-4 consisting of 868 RGB sequential image frames. Figure \ref{fig:poseplots}(a-c) illustrates the ground truth and estimated pose-trajectories for: 4-bit classical (deterministic) inference and (MC-dropout-based) 4-bit probabilistic inference with 30 samples per image frame. Figure \ref{fig:poseplots}(d) depicts a scatter plot of pose-error vs. variance (uncertainty) in the probabilistic inference. \textit{Interestingly}, a high correlation between the error and predictive uncertainty can be seen. Therefore, unlike classical (deterministic) inference, MC-CIM's probabilistic inference can indicate when the mispredictions are likely by expressing high predictive uncertainties. The quantitative estimate of error-uncertainty correlation (Pearson correlation coefficient \cite{lee1988thirteen}) observed in Figure \ref{fig:poseplots}(d) is 0.31. 

In Figure \ref{fig:poseplots}(e-f), we study the impact of MC-CIM's non-idealities on error-variance correlation. In Figure \ref{fig:poseplots}(e), we observe a good error-uncertainty correlation (> 0.3) at 4-bit input/weight precision onward in MC-CIM's probabilistic inference. The implication of dropout probability ($p_1$)'s bias perturbation due to CIM-embedded RNG's non-idealities in MC-Dropout inference is shown in Figure \ref{fig:poseplots}(f). On the x-axis is dropout probability sampled from symmetric Beta distribution, $p \sim\mathcal{B}(a,a)$, with varying degrees of variance parametrized by $a$. The error-uncertainty correlation in pose estimates is reasonably good even at high dropout probability perturbations where $a$ is as low as 2 shown in Figure \ref{fig:poseplots}(f). The correlation drops for $p1 \sim\mathcal{B}(1.25,1.25)$ as shown in the Figure. Thus MC-CIM proves to be robust amidst its non-idealities in performing confidence-aware self-localization. 

\section{Conclusion}
We presented MC-CIM, a compute-in-memory framework for probabilistic inference targeting edge platforms that not only gives prediction but also the confidence of prediction which is crucial for risk-aware applications such as drone autonomy and augmented/virtual reality. For Monte Carlo Dropout (MC-Dropout)-based probabilistic inference, MC-CIM is embedded with dropout bits generation and optimized computing flow to minimize the workload and data movements. With our proposed techniques we benefit significantly in energy savings even with additional probabilistic primitives in CIM framework. Our study on the implications on non-idealities in MC-CIM on probabilistic inference shows promising robustness of the framework for two applications - mis-oriented handwritten digit recognition and confidence-aware visual odometry in drones.

\vspace{5mm}
\noindent \textbf{Acknowledgement:} This work was supported by NSF CAREER Award (2046435) and a gift funding from Intel.

\bibliographystyle{IEEEtran}
\bibliography{main.bib}
\end{document}